\documentclass[table,xcdraw]{article} 
\usepackage{iclr2021_conference,times}
\usepackage{graphicx}
\usepackage{booktabs}

\usepackage{amsmath,amsfonts,bm}

\def\eqref#1{equation~\ref{#1}}

\def\1{\bm{1}}

\def\mX{{\bm{X}}}

\DeclareMathAlphabet{\mathsfit}{\encodingdefault}{\sfdefault}{m}{sl}
\SetMathAlphabet{\mathsfit}{bold}{\encodingdefault}{\sfdefault}{bx}{n}

\usepackage{hyperref}
\usepackage{url}

\title{MONCAE: Multi-Objective Neuroevolution of Convolutional Autoencoders}

\author{Daniel Dimanov, Emili Balaguer-Ballester\\
Faculty of Science and Technology, Bournemouth University\\
Bournemouth, BH12 5BB, UK \\
\texttt{\{ddimanov,eb-ballester\}@bournemouth.ac.uk} \\
\And
Colin Singleton \\
CountingLab \\
Reading, UK \\
\texttt{colin@countinglab.co.uk}
\And
Shahin Rostami\\
Data Science Lab, Polyra Limited\\
Bournemouth, UK\\
\texttt{shahin@polyra.com}
}

\iclrfinalcopy 
\begin{document}

\maketitle

\begin{abstract}
In this paper, we present a novel neuroevolutionary method to identify the architecture and hyperparameters of convolutional autoencoders. Remarkably, we used a hypervolume indicator in the context of neural architecture search for autoencoders, for the first time to our current knowledge. Results show that images were compressed by a factor of more than 10, while still retaining enough information to achieve image classification for the majority of the tasks. Thus, this new approach can be used to speed up the AutoML pipeline for image compression.

\end{abstract}
\section{Introduction}

Convolutional neural networks have achieved exceptional results for computer vision problems \citep{goodfellow2016deeplearningbook}, but they require expertise to be constructed. Thus, until recently, researchers relied mainly on trial and error to determine the best hyperparameters and network architecture \citep{stanley2019designing}. Recently, there has been an increase in the number of effective approaches for automated convolutional network design, such as evolutionary algorithms \citep{lu2019nsga,miikkulainen2019evolving,real2019regularized,stanley2019designing}, reinforcement learning \citep{qin2019nasnet,tan2019efficientnet} and other automated approaches \citep{he2021automl}.These methods have been able to match and sometimes surpass expert-informed, manually designed architectures.

One of the most successful approaches for automated machine learning has been neuroevolution (NE) \citep{real2019regularized,lu2019nsga}, but its long training times makes it unfeasible for a plethora of problems \citep{real2019regularized}.
Interestingly, it was recently shown that choosing a curated set of attributes or building new ones from the original features, can reduce the computational cost and even improve the performance \citep{charte2020evoaaa}.To overcome this drawback, this paper proposes a neural architecture search approach based on neuroevolution to approximate the Pareto-front of convolutional autoencoders. This method is designed to optimize the trade-off between reconstruction loss and image compression (calculated based on the size of the bottleneck layer). To the best of the authors' knowledge, this is the first attempt to do neural architecture search for convolutional autoencoders and to use multi-objective optimisation technique in this context.

\section{Related work}
\label{rel-work}

\paragraph{Autoencoders}
In recent years, autoencoder neural networks have shown promise for addressing a range of supervised and unsupervised problems \citep{ballard1987modular,schmidhuber2015deep}. Autoencoders are NNs with symmetric input and output layers, termed encoder and decoder, that usually contain a bottleneck layer in between \citep{schmidhuber2015deep,goodfellow2016deeplearningbook}. The task they solve is to minimise the reconstruction loss, which is calculated based on the difference between the original input and the reconstruction (output) \citep{baldi2012autoencoders}. This allows for various applications ranging from style transfer \citep{qian2019autovc} and generative techniques \citep{dosovitskiy2016generating} to semi-supervised learning \citep{akcay2018ganomaly}, data denoising \citep{gondara2016medical} and image compression techniques \citep{charte2020evoaaa} 
For more sophisticated datasets, convolutional layers can be incorporated instead of simple dense layers, these autoencoders are referred to as convolutional autoencoders \citep{azarang2019convolutional}. In difference with simple autoencoders, they achieve the compression through the use of convolutional and pooling layers instead of simply having fewer neurons and the upsampling is achieved through deconvolutions (also known as upsampling layers) \citep{baldi2012autoencoders}. The state-of-the-art in autoencoders is composed of architectures designed by experts, which are intended to work for the specific scenarios \citep{charte2020evoaaa}. Currently, given the increasing interest on Automated machine learning(AutoML) \citep{he2021automl}, a range of automated approaches for designing and optimisation architectures have been proposed \citep{real2019regularized,lu2019nsga,yan2020does}.

\paragraph{Neural Architecture Search, AutoML and Neuroevolution}

The rise in popularity of AutoML is s natural response to address the high demand for expertise and computational resources for deep learning in industry and academia \citep{he2021automl,stanley2019designing}. The automation involves not the NN training, but also the design of architectures, which is known as Neural Architecture Search (NAS) \citep{stanley2019designing,elsken2019neural}.
Multiple different algorithms and approaches have been proposed with the first significant one for modern NAS being the work of \citep{zoph2016neural}. As the foundations were laid, various reinforcement learning \citep{qin2019nasnet,tan2019efficientnet}, Neuroevolution \citep{lu2019nsga,miikkulainen2019evolving,real2019regularized,stanley2019designing} and other approaches started to emerge. 

A significant breakthrough in the field occurred in 2017, when \citet{real2019regularized}'s AmoebaNET reached state-of-the-art performance, outperforming architectures designed by experts. The AmoebaNet is a Neuroevolution algorithm \citep{real2019regularized}. In such approaches, decoding the NN (the individual in an evolutionary algorithm context) is added as an extra step before the evaluation, whilst encoding the individual back to genotype is added after the evaluation \citep{stanley2019designing}. Recently, indirect encodings \citep{hadjiivanov2016complexity} (closely inspired by how DNA works \citep{miikkulainen2019evolving}) have shown promising results. 
Neuroevolution has been successfully used for autoencoder optimization in a range of studies \citep{charte2020evoaaa,sereno2018automatic,alvernaz2017autoencoder}; however, a limitation of these approaches is that they typically focus on evolving a NN with dense activation layers (e.g., in \citet{charte2020evoaaa}'s EvoAAA) instead of convolutional and pooling layers, which have shown great potential for computer vision \citep{gu2018recent}.

\section{MONCAE}
\label{MOOConvAENE}
MONCAE \footnote{*URL will be provided for the camera-ready version} is built on top of DEvol \citep{devol}\footnote{https://github.com/joeddav/devol}, which is a neuroevolution algorithm for image classification. We modified this algorithm to allow for searching optimal convolutional autoencoders, and to evaluate the population using multi-objective optimisation. Thus, to the best of our knowledge, this is the first attempt of using neuroevolution for automating the design of convolutional autoencoders; whilst incorporating multi-objective optimisation for their automated NAS. As we use DEvol as the backbone of this approach the evolutionary algorithm behind it is a genetic algorithm, which uses tournament selection and both crossover and mutation as variation operators.

A severe limitation of AutoML approaches (and neuroevolution in particular) is that these algorithms take a tremendous amount of time to discover optimal solutions \citep{he2021automl}. In the case of AmoebaNET, the time exceeds 50000 GPU hours for grayscaled CIFAR-10 (Appendix \ref{apex:data}),

which makes it unfeasible for many modern-day scenarios \citep{lebedev2018speeding}. There is a plethora of different attempts for solving this problem, focusing both in the 'data space' \citep{wang2018dataset,singh2017hide} as well as in the 'algorithmic space' \citep{hinton2015distilling,frankle2018lottery}. Recently, researchers have demonstrated that one promising approach is to split the problem into two stages: first feature engineering and then using the extracted features to solve the problem at hand \citep{sereno2018automatic}. To achieve that, \citet{sereno2018automatic} and \citet{charte2020evoaaa} have showcased the possibility of using an automatically discovered autoencoder to tackle this. 

We followed a similar methodology, but instead of using dense layers for our architecture, we allowed the neuroevolution algorithm to search for convolutional and pooling layers too. Moreover, we implemented the hypervolume indicator~(HVI) (Equation \ref{eq:hvi}) \citep{rostami2016covariance,guerreiro2020hypervolume} to find a trade-off between two main objectives: 1) the level of compression, for which we use a metric, which is roughly based on Bayesian information criterion \citep{watanabe2013widely} (Equation \ref{eq:loc}) and 2) the reconstruction loss. 

Remarkably, with our approach, further objectives can be added. In Equation \ref{eq:hvi},  $\mX$ refers to the matrix containing all performance metrics for the whole population set, $\textbf{r}$ refers to the vector of reference points for each objective ($f^{ref}$ per objective) , $m$ refers to the number of objectives, and $f_{m}$ refers to the objective function for objective $m$.

\begin{equation}
    \label{eq:loc}
    Level~of~compression(\textbf{k}) =  log_{10}(\prod_{i=0}^{dimensions}(\textbf{k}_{i}))
\end{equation}

\begin{equation}
    \label{eq:hvi}
    HVI(\textbf{X},\textbf{r}) = \Lambda \left ( \bigcup_{\textbf{x}_{n} \in X} [f_{1}(\textbf{x}_{n},f_{1}^{ref})]  \cdots  [f_{m}(\textbf{x}_{n},f_{m}^{ref})] \right ) 
\end{equation}
Based on these two objectives and prespecified reference points for each one, we first calculated the HVI of the population at each generation, and then evaluated the contribution of each individual to the overall HVI (CHVI, Equation \ref{eq:chvi}), which used as the fitness score. The process follows the general Neuroevolution cycle; details for each step can be found in Appendix \ref{apex:steps}.

\begin{equation}
    \label{eq:chvi}
    CHVI_{\textbf{x}_{i}} = HVI(\textbf{X} \cup \{\textbf{x}_{i}\}) -  HVI(\textbf{X} \setminus  \textbf{x}_{i})
\end{equation}

\paragraph{Experimental Setup}
We evaluate MONCAE on three datasets: MNIST \citep{lecun1999mnist}, Fashion-MNIST \citep{xiao2017fashion} and CIFAR-10 \citep{krizhevsky2014cifar} (see dataset details in Appendix \ref{apex:data}). For each dataset, we do 10 runs with different random seeds (Appendix \ref{apex:seeds}) for 20 generations and a population size of 20, a maximum of 5 epochs and using early stopping. After the process is complete, we finetune the final population by training for an extra 20 epochs using the parameters, hyperparameters, optimiser and architecture discovered by the algorithm. Experiments were implemented in python, TensorFlow and a single RTX 3090 NVidia GPU with CUDA 11.1.  

Using multi-objective optimisation with a population-based approach such as neuroevolution allows us to examine the whole final approximation set of produced solutions, instead of picking one. This can allow the stakeholders to choose the model that best fits the context of the problem at hand. Moreover, setting the reference points can also induce bias towards certain solutions and enable preference articulation to be added to the optimisation process. 

\section{Results}
\label{sec:res}

Average results over the whole population would undermine our purpose of producing an approximation set from which to choose. Thus, we present the results from the runs having latent representations which compress the input by at least a factor of 2, since more accurate, but larger representations are not the focus of this work. The results in  Table \ref{tab:results-overview} show averaged results for models that have a level of compression below a certain threshold (2.5 for MNIST and F-MNIST and 3.1 for CIFAR-10) in the final population.

\begin{figure}[h]
\begin{center}
\includegraphics[width=0.5\linewidth]{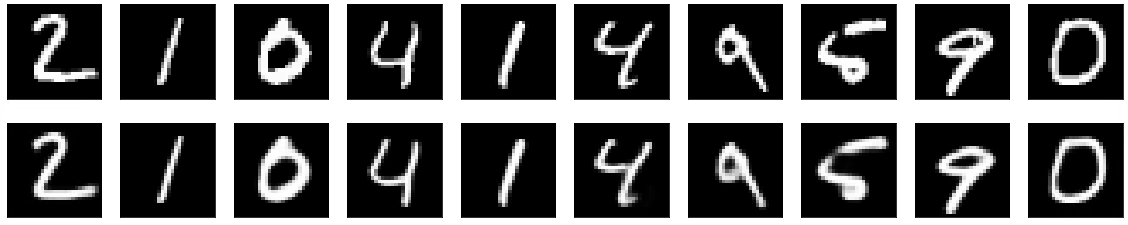}
\end{center}
\caption{MNIST autoencoder with bottleneck of 4x4x4 and reconstruction loss of 0.0817}
\label{fig:mnist-best}
\end{figure}

\begin{figure}[!h]
\begin{center}
\includegraphics[width=0.49\linewidth]{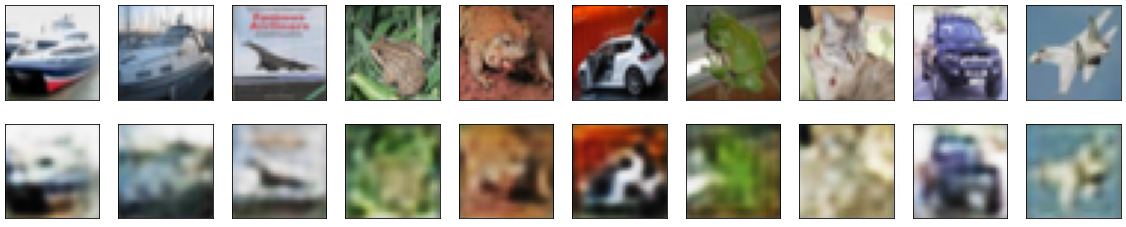}
\includegraphics[width=0.49\linewidth]{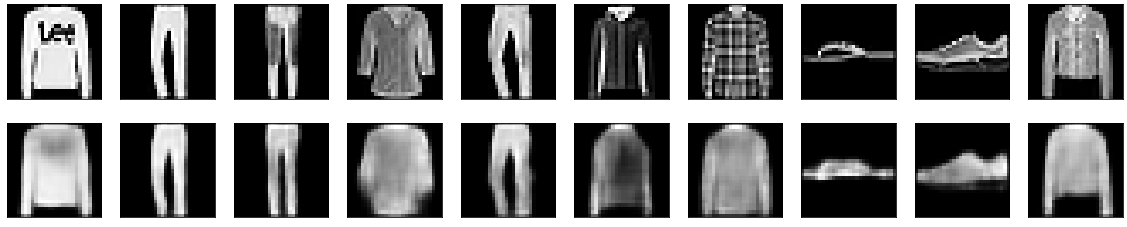}

\end{center}
\caption{Left: Fashion-MNIST autoencoder with bottleneck of 4x4x4 and reconstruction loss of 0.289. Right: CIFAR-10 autoencoder with bottleneck of 8x8x8 and reconstruction loss of 0.564}
\label{fig:cifar10-best}
\end{figure}

A sample population from one of the runs can be seen it Appendix \ref{apex:sample-pop}, which showcases the variety of available models produced. The total time in the table is the time for the whole algorithm to do 1 run with the constraints specified above. As expected, the reconstruction and classification error is the lowest for MNIST. The average level of compression is around 1.7 (approximately a 4x4x3 representation), which means that the autoencoder managed to compress the data by a factor of almost \textbf{16} on average. The 85.8\% accuracy for MNIST is not state-of-the-art level, but, remarkably, these architectures were discovered using only 20 generations with a population size of 20 for around 1.5 GPU hours on average. A more detailed look at one of the best performing models for MNIST is available in Figure \ref{fig:mnist-best}. The same can be found for F-MNIST and CIFAR-10 in Figure \ref{fig:cifar10-best}. 

\begin{table}[ht]
\caption{Summarised results averaged over 10 runs and all individuals with compression below 3.1 for CIFAR-10 and 2.5 for MNIST and F-MNIST dasets.}
\label{tab:results-overview}
\begin{center}
\small
\begin{tabular}{@{}ccccccc@{}}
 & \textbf{Rec loss} & \textbf{LOC} & \textbf{Cl loss} & \textbf{Cl acc} & \textbf{\begin{tabular}[c]{@{}c@{}}Total time\\ w/o ft (min)\end{tabular}} & \textbf{\begin{tabular}[c]{@{}c@{}}Finetuning\\ time (s)\end{tabular}} \\ \midrule
\rowcolor[HTML]{C0C0C0} 
\multicolumn{1}{c|}{\cellcolor[HTML]{C0C0C0}\textbf{MNIST}} & 0.141 & 1.699 & 0.409 & 0.858 & 93.23 & 1160 \\
\multicolumn{1}{c|}{\textbf{F-MNIST}} & \cellcolor[HTML]{FFFFFF}0.316 & \cellcolor[HTML]{FFFFFF}1.639 & \cellcolor[HTML]{FFFFFF}0.524 & \cellcolor[HTML]{FFFFFF}0.803 & 91.19 & 889 \\
\rowcolor[HTML]{C0C0C0} 
\multicolumn{1}{c|}{\cellcolor[HTML]{C0C0C0}\textbf{CIFAR-10}} & 0.596 & 2.15 & 1.693 & 0.402 & 123.04 & 2752
\end{tabular}
\end{center}
\end{table}

However, the fact that the algorithm managed to compress the dataset to representations that are (2x2x4) while still retaining around 50\% of the data, shows that the approach is indeed promising. With larger population sizes, more epochs, more generations and expanding the search space, this approach can be scaled to larger datasets. Nevertheless, to scale the approach to multi-channel inputs remains a challenge (Figure \ref{fig:cifar10-best}).

Figures \ref{fig:mnist-best} and \ref{fig:cifar10-best} present some of the best results obtained for all of the three datasets, showcasing the balance between compression and reconstruction loss. From the images, it can be seen that the MNIST reconstructions are hardly distinguishable from the originals. The results for F-MNIST are also still recognisable, but most of the details are lost during compression. Images for CIFAR-10, on the other hand, barely resemble the original images for the most part, which illustrates how adding the extra colour channels and complex objects increases the complexity of the problem, and hence limits the extent to which a dataset can be compressed in lower-dimensional representations.

Noteworthy, some of the objects in the last two images of \ref{fig:cifar10-best} look similar. 
In the presented model the algorithm managed to compress the representations of these images by a factor of 12 for MNIST (Figure \ref{fig:mnist-best}) and F-MNIST (Figure \ref{fig:cifar10-best}), while all useful information (except the sandal class for F-MNIST) seems to have been retained. As per CIFAR-10, the images were decreased by a factor of 6, while retaining some of the information, but classification performance dropped excessively.

\section{Conclusion and Future Work}
This paper presents a novel approach for automatically constructing convolutional autoencoders using neural architecture search, which utilizes neuroevolution to approximate the Pareto-front of solutions. While some of our results, especially for CIFAR-10, are still sub-optimal, this study is designed to be a stepping stone towards automating the process and making it considerably faster. Our goal is to speed up AutoML and allow decision-makers to articulate preference, offering them a set of models to choose from, instead of just a single one. We believe that the generations and population size we chose are insufficient to explore such an enormous search space. In future studies, we aim to scale the algorithm for significantly larger datasets and expand the search space available by adding new hyperparameters, as well as potentially achieving end-to-end neuroevolution. We are also interested in adding extra objectives in our multi-objective optimisation (e.g., explainability).



\bibliography{iclr2021_conference}
\bibliographystyle{iclr2021_conference}

\appendix
\section{Datasets}
\label{apex:data}
\paragraph{MNIST}
MNIST is a dataset containing 70000 images of handwritten digits in total, which are split among 10 classes. Each class represents an Arabic digit (0-9) and the images are 28x28 pixels and only have 1 colour channel, where the white value is specified as pixel intensity between 0-255 \citep{lecun1999mnist}.

\paragraph{F-MNIST}
Fashion MNIST is a similar dataset to MNIST with the same number of images as well as image shape, but the only difference is that the 10 classes are not digits, but instead 10 different pieces of clothing \citep{xiao2017fashion}. 

\paragraph{CIFAR}
CIFAR is a labelled subset of the Tiny Images Dataset \citep{prabhu2020large}. There are two main versions of the CIFAR dataset and these are CIFAR-10 (with 10 classes) and CIFAR-100 (with 100 classes) \citep{krizhevsky2014cifar}. The main difference with MNIST and F-MNIST is that this dataset is severely more complex and not only the images are 32x32, but they also have 3 (RGB) channels, so the total pixel values are 3072, compared to the 784 of the other two datasets.

\section{MONCAE Neuroevolution step details}
\label{apex:steps}

\begin{figure}[h]
\begin{center}
\includegraphics[width=0.8\linewidth]{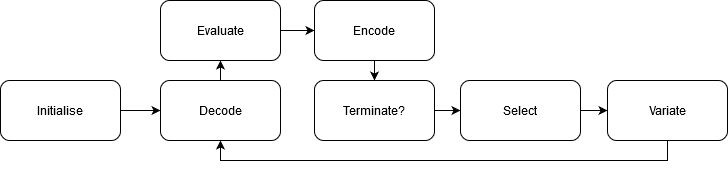}
\end{center}
\caption{High-level flowchart of Neuroevolution with all steps from general evolutionary algorithms with Decode and Encode operators added. The Variation operator can be further split into Crossover and Mutation in the case of Genetic algorithms.}
\label{fig:ne}
\end{figure}

First, some algorithm hyperparameters are set. These are the number of generations, the population size, maximum number of convolutional (and pooling) layers, maximum filter size in a convolutional layer and the maximum epochs (used when each architecture is evaluated). The general steps of the algorithm follow the neuroevolution process depicted in Figure \ref{fig:ne}. 

For the \textbf{initialisation} we create a random population (of the specified size). All individuals are by default \textbf{encoded} in a conveniently created python class, which stores the genome, which is composed of a long 2-dimensional list, which contains the information about all layers of the model (including a boolean value specifying if the layer is turned on or off, how many nodes/filters are there in the layer, what is the type of layer (convolutional, pooling, dense etc.), what activation function is the layer using, is there a batch normalisation following the layer, dropout etc.) as well as other hyperparameters, which include the optimiser the model is using and the learning rate. 
After the population is initialised, the individuals are \textbf{decoded} into TensorFlow models and are further \textbf{evaluated} by training for 5 epochs and then verifying the results by running the model on a pre-defined validation set that all individuals use. Here we tried various ways of reporting the fitness of each individual, but since we were aiming to minimise the latent vectors in the bottleneck as well as minimise reconstruction loss we chose to use the hypervolume indicator as a metric with the contributing hypervolume of each individual being its fitness score. So, the first objective is straightforward, which is the reconstruction loss, but for the size of the latent representations, we chose to name this "level of compression" and the way we calculate the level of compression is roughly based on Bayesian information criterion \citep{watanabe2013widely} (Equation \ref{eq:loc}). The \textbf{l} represent the latent representation as each element in the vector is the size of a dimension.

The next step, (\textbf{Terminate}), simply checks if the termination criterion is met and if it is, then the algorithm is stopped (in the context of these experiments, this criteria is the number of generations). For the \textbf{selection} a certain per cent of the best individuals survive and are kept in the next generation. For other individuals to be selected they have to have a higher fitness score than a random score (which has a bound of 0 and the maximum score for the generation). The last step, before the cycle repeats, is to \textbf{variate} and the population and for variation, this approach uses both crossover and mutation, which is done on the selected individuals to form the next generation.

In our experiments, the reference points are set to 4 and 12 for the error and level of compression correspondingly. Thus, the algorithm might be biased to pick models with better compression than ones with lower error. In all training stages, if not otherwise specified, we used a batch size of 256 and trained for 20 epochs.

\section{Sample population for MNIST}
\label{apex:sample-pop}
\begin{figure}[h]
\begin{center}
\includegraphics[width=0.7\linewidth]{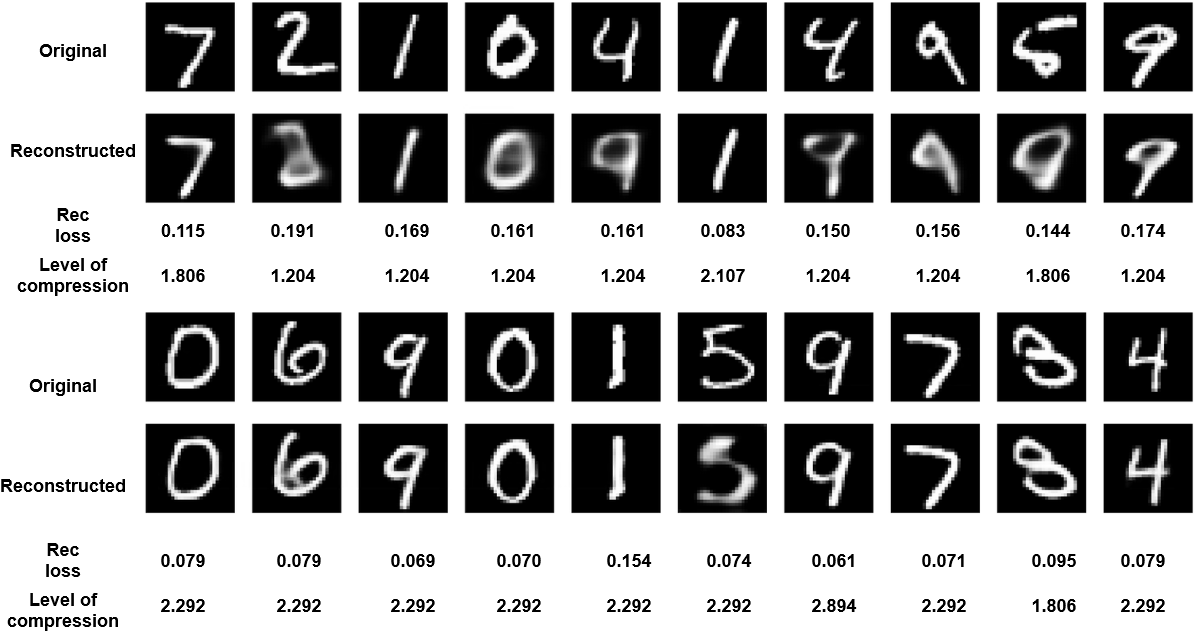}
\end{center}
\caption{A population of solutions for MNIST. A single digit is used as example of the performance of each single model from a run picked from the authors. For each model an example of reconstruction of a single digit can be seen as well as the achived reconstruction loss and level of compression.}
\label{fig:pop-mnist}
\end{figure}

\section{Used seeds}
\label{apex:seeds}
[46, 29, 12, 8, 68, 44, 32, 91, 85, 61]

\end{document}